\pdfoutput=1

\documentclass[11pt]{article}

\usepackage{natbib}
\usepackage{hyperref}
\usepackage[final]{acl}
\usepackage{booktabs}
\usepackage{multirow}
\usepackage{times}
\usepackage{latexsym}
\usepackage{tabularx} 
\usepackage{geometry}
\usepackage{graphicx}
\usepackage[T1]{fontenc}

\usepackage[utf8]{inputenc}

\usepackage{microtype}

\usepackage{inconsolata}

\usepackage[framemethod=TikZ]{mdframed}
\usepackage{lipsum} 

\mdfdefinestyle{exampledefault}{
  linecolor=black,
  outerlinewidth=2pt,
  roundcorner=20pt,
  innertopmargin=10pt,
  innerbottommargin=10pt,
  innerrightmargin=10pt,
  innerleftmargin=10pt,
  backgroundcolor=white
}

\title{Syn-QA2: Evaluating False Assumptions in\\Long-tail Questions with Synthetic QA Datasets}

\author{Ashwin Daswani \hspace{0.2cm} Rohan Sawant \hspace{0.2cm} Najoung Kim\\
  Boston University\\
  \texttt{\{ashwind,rohan16,najoung\}@bu.edu }\\}

\begin{document}
\maketitle
\author{}

\begin{abstract}
Sensitivity to false assumptions (or false premises) in information-seeking questions is critical for robust question-answering (QA) systems. Recent work has shown that false assumptions in naturally occurring questions pose challenges to current models, with low performance on both generative QA and simple detection tasks \citep{kim-etal-2023-qa}. However, the focus of existing work on naturally occurring questions leads to a gap in the analysis of model behavior on the long tail of the distribution of possible questions. To this end, we introduce Syn-(QA)$^2$, a set of two synthetically generated QA datasets: one generated using perturbed relations from Wikidata, and the other by perturbing HotpotQA \citep{yang-etal-2018-hotpotqa}. Our findings from evaluating a range of large language models are threefold: (1) false assumptions in QA are challenging, echoing the findings of prior work, (2) the binary detection task is challenging even compared to the difficulty of generative QA itself, possibly due to the linguistic structure of the problem, and (3) the detection task is more challenging with long-tail questions compared to naturally occurring questions, highlighting the utility of our synthetic datasets and generation method.
\end{abstract}

\section{Motivation}
Information-seeking questions with false assumptions like \textit{When did Mark Zuckerberg found Google?} pose challenges to question-answering (QA) systems because they require questioning the correctness of the underlying assumptions, or presuppositions of the questions (i.e., systems must recognize that \textit{Mark Zuckerberg founded Google} is false). Recent work has shown that a wide range of QA systems, including systems based on large language models (LLMs), do struggle with such questions \citep{kim-etal-2023-qa,yu-etal-2023-crepe,hu-etal-2023-wont,vu2023freshllms}. However, existing work has primarily focused on naturally occurring questions---the questions were sourced from popular queries to a search engine \citep{kim-etal-2021-linguist,kim-etal-2023-qa}, Reddit posts \citep{yu-etal-2023-crepe}, or were human-written \citep{hu-etal-2023-wont,vu2023freshllms}. There are two limitations to this approach: first, this leads to a gap in the analysis of model behavior on the long tail of the distribution of possible questions, and second, the evaluation data tend to be small due to annotation efforts required. For example, the largest test set among mentioned work is from \citealt{yu-etal-2023-crepe}, with 751 test instances of questions with false assumptions. 

\begin{figure}[t]
\centering
\includegraphics[width=0.37\textwidth]{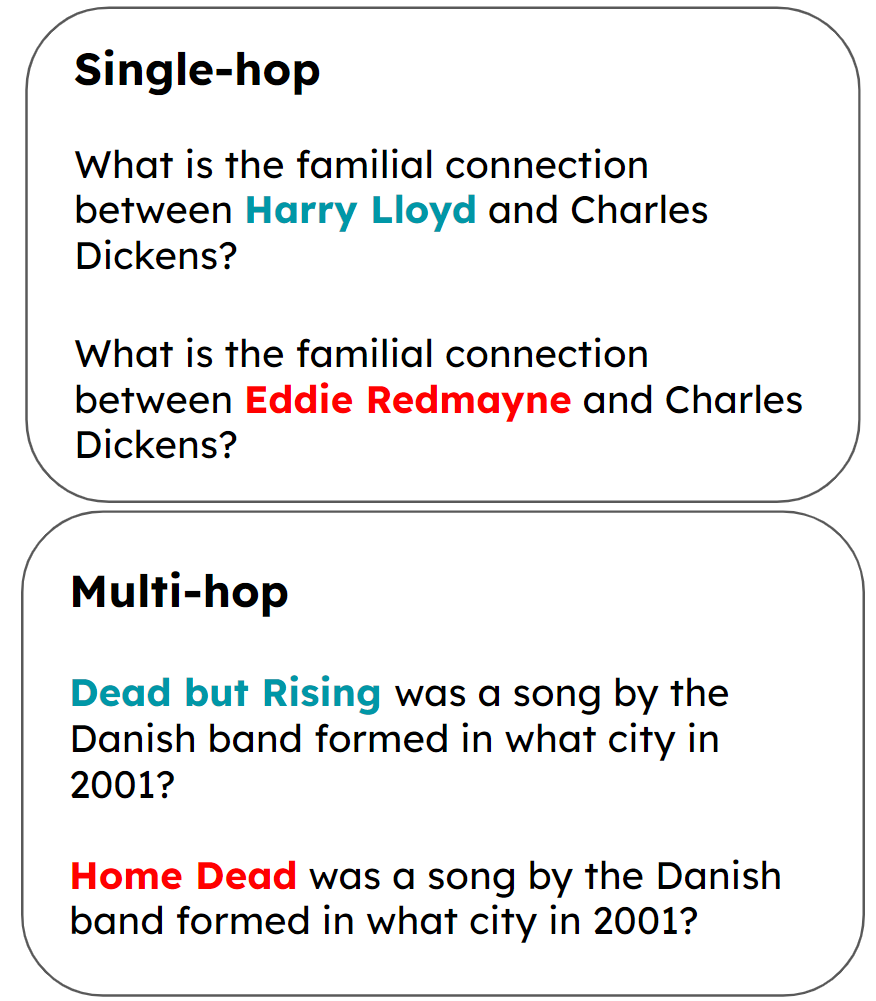}
\caption{Example of a minimal pair of questions without and with false assumptions in our single-hop and multi-hop datasets. The perturbed entity that is responsible for the false assumption is colored \textbf{\textcolor{red}{red}} and the pre-perturbation entity is colored \textbf{\textcolor{teal}{teal}}.}
\label{fig:examples}
\end{figure}

\begin{figure*}[t]
\centering
\includegraphics[width=0.8\textwidth]{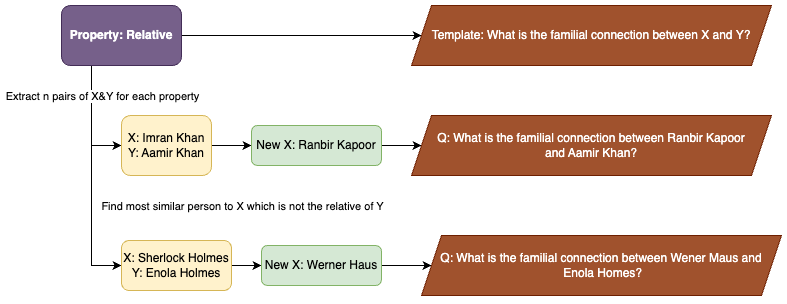}
\caption{Visualization of the single-hop question generation process.}
\label{fig:singlehopexample}
\end{figure*}

To this end, we introduce Syn-(QA)$^2$ (synthetic questions with questionable assumptions, inspired by (QA)$^2$ of \citealt{kim-etal-2023-qa}),\footnote{The dataset is available at \url{https://github.com/ashwindaswanibu/QAQA-Synthetic-Dataset}.} a collection of two synthetic English QA datasets generated using entity perturbation on \href{https://www.wikidata.org}{Wikidata} and HotpotQA \citep{yang-etal-2018-hotpotqa}, to evaluate the effect of false assumptions in both single-hop and multi-hop QA. The perturbation approach we adopt is similar to the entity replacement method of \citet{gautam-etal-2023-lightweight}, but our main focus lies on open-domain questions rather than on reading comprehension questions. Furthermore, this approach enables a minimal pair comparison between questions with and without false assumptions (see Figure~\ref{fig:examples}), allowing us to quantify the effect of false assumptions without the effect of additional variations such as the phrasing and style of the questions. 

We evaluate multiple LLMs on Syn-(QA)$^2$ and find that: (1) false assumptions still pose substantial challenges to current models, corroborating recent findings, (2) binary false assumption detection is challenging even compared to the difficulty of generative QA itself, possibly due to the linguistic structure of the problem: it embeds a question inside a question, and (3) the detection task is more challenging with long-tail questions compared to naturally occurring questions, highlighting the utility of our synthetic datasets and generation method.

\begin{figure*}[t]
\centering
\includegraphics[width=1\textwidth]{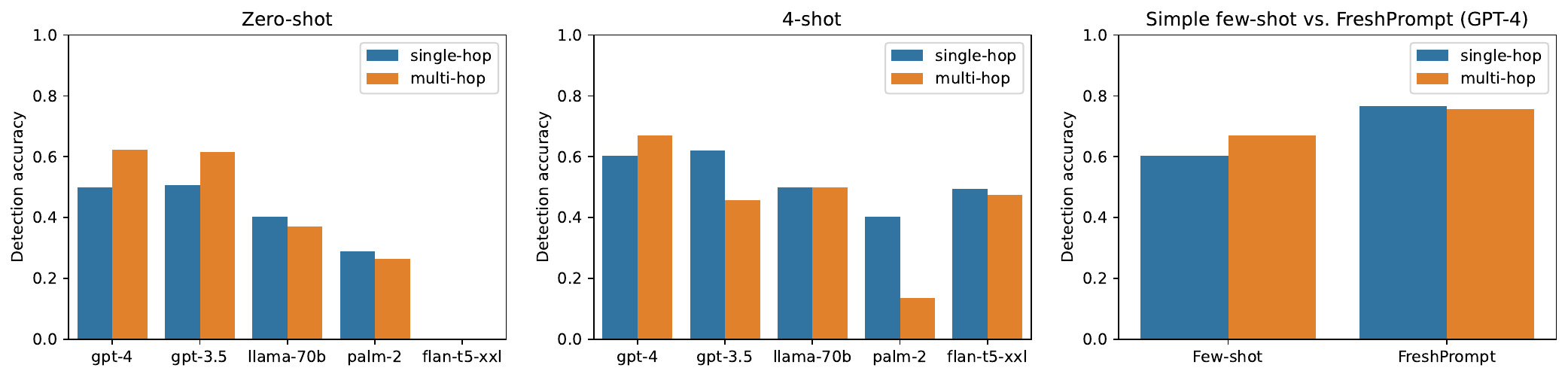}
\caption{Accuracy for the false assumption detection task on our datasets. Since few-shot and few-shot CoT did not show substantially different trends, we only show the few-shot results here (see Appendix~\ref{app:full-results} for full results). Zero-shot Flan-T5-XXL did not yield correct class labels in most cases.}
\label{fig:results-detection}
\end{figure*}

\section{Dataset}
Syn-(QA)$^2$ contains 1812 minimal pairs of questions in total (1165 single-hop, 647 multi-hop); each pair consists of questions with and without false assumptions, differing only in terms of a single entity mention.  We detail the generation process for each dataset below.

\subsection{Generating single-hop questions with false assumptions from Wikidata relations}
\label{subsec:gen-single-hop}
We generated single-hop questions containing false assumptions from Wikidata relations through the following process, also visualized in Figure~\ref{fig:singlehopexample}.

\begin{enumerate}
    \item \textbf{Relation triple sampling}: Select a set of relations \texttt{rel} that hold between two entities \texttt{x} and \texttt{y} from Wikidata (e.g., \texttt{relative-of}, \texttt{educated-at}). Sample multiple entity pairs \texttt{(x, y)} for each \texttt{rel} to create a set of true relation triples (e.g., \texttt{(relative-of, Imran Khan, Aamir Khan)}). \vspace{-0.2cm}
    \item \textbf{Similarity-based entity replacement}: For each relation triple, substitute \texttt{x} with a different entity \texttt{x$'$} that has high similarity with \texttt{x} based on the number overlapping Wikidata properties (e.g., \texttt{Imran Khan} $\rightarrow$ \texttt{Ranbir Kapoor}; see Appendix~\ref{app:properties} for a full list of properties used). Skip replacement if there is no sufficiently similar entity based on a threshold value $\theta$ (we used $\theta=3$). \vspace{-0.2cm}
    \item \textbf{Template population}: Write natural language templates of \textit{wh-}questions embedding the assumption that \texttt{(rel,x,y)} holds. For example, \textit{What is the familial connection between x and y?} assumes that there is a familial connection between \texttt{x} and \texttt{y}---i.e., \texttt{(relative-of,x,y)} is true. Populate the templates with \texttt{(rel,x,y)} and \texttt{(rel,x$'$,y)} to create minimal pairs of questions varying in whether they contain a false assumption. \vspace{-0.2cm}
    \item \textbf{Manual verification}: Remove question pairs where \texttt{(rel,x$'$,y)} in fact holds. 
\end{enumerate}

\noindent We selected 21 relations and sampled 200 distinct entity pairs for each triple, yielding a set of 1953 candidate questions (this count is lower than \# rel $\times$ \# samples because of the similarity thresholding in Step 2). 1165 question pairs remained after manually removing cases where false assumptions were not introduced by entity replacement.

\subsection{Generating multi-hop questions with false assumptions from HotpotQA}
We generated multi-hop questions containing false assumptions from HotpotQA using the distractor information already included in the dev-distractor set of the original dataset. This set contains distractor documents in addition to true supporting documents relevant to the original question. We replaced the title of the true supporting document (if the string exists in the question) with the title of the most similar distractor document. The similarity is computed based on the number of shared Wikipedia categories between the pages of the original entity and the distractor entity. For example, the Wikipedia pages of \texttt{Roger O. Egeberg} and \texttt{Steven K. Galson} share categories like \textit{American public health doctors} and \textit{Articles with ISNI identifiers}. After this similarity-based replacement, we manually verified the questions and removed nonsensical questions or questions without false assumptions. The initial replacement process yielded 893 pairs of multi-hop questions with and without false assumptions, and we included 647 question pairs in the final dataset after manual verification.

\begin{figure*}[t]
\centering
\includegraphics[width=0.84\textwidth]{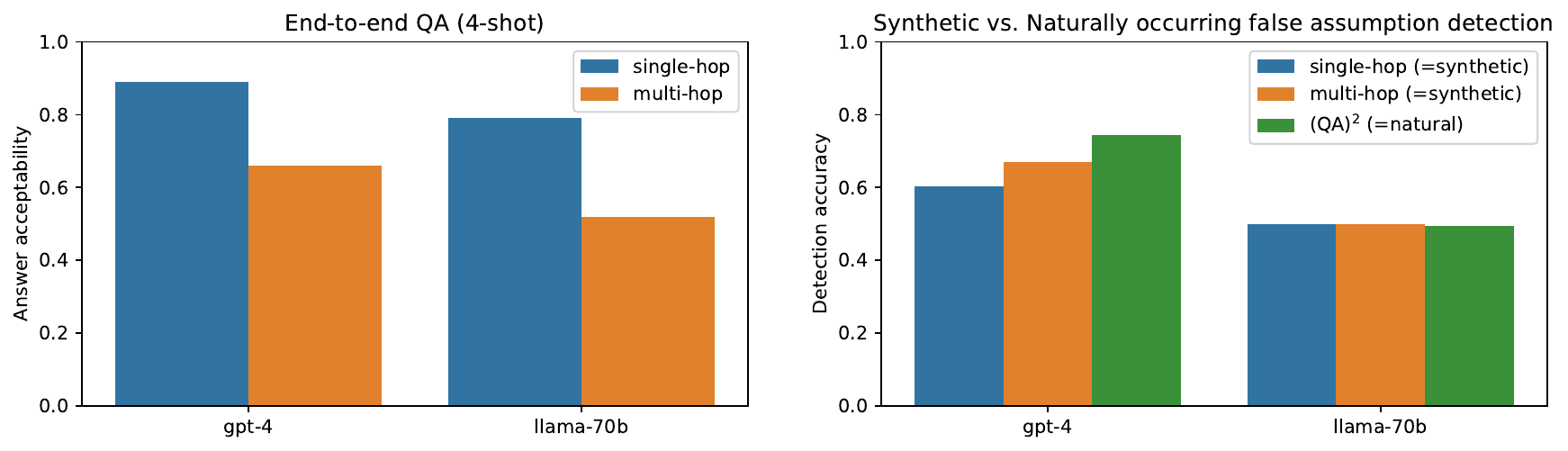}
\caption{Results of the generative QA evaluation (left) and results comparing the difficulty of false assumption detection for synthetic and naturally occurring questions (right).}
\label{fig:results-misc}
\end{figure*}

\section{Experiments}

\subsection{Evaluation metrics}
Our main automatic metric is the accuracy on the false assumption detection task proposed in \citet{kim-etal-2023-qa}, which has been shown to correlate well with human acceptability on actual generative QA. False assumption detection is a binary classification task of identifying whether the input question contains a false assumption or not (e.g., \textit{Does the question ``What is the familial connection between Ranbir Kapoor and Aamir Khan'' contain a false assumption, Yes or No?}).

We additionally conducted manual evaluation of generative QA performance, measured by binary acceptability of the model answer annotated by two of the authors, again following \citet{kim-etal-2023-qa}.

\subsection{Models and Prompting}
We evaluated a range of LLMs: GPT-3.5 (\texttt{gpt-3.5-turbo}), GPT-4 \citep{achiam2023gpt}, Llama-2-70B (4-bit quantized) \citep{touvron2023llama}, PaLM-2 (\citealt{anil2023palm}; \texttt{text-bison-001}), and Flan-T5-XXL \citep{chung2022scaling}, under zero-shot, few-shot ($k=4$, two questions with and without false assumptions each), and few-shot with chain-of-thought settings. We also evaluated GPT-4 combined with FreshPrompt \citep{vu2023freshllms}, a prompting strategy that leverages search-engine retrieved contexts, on a subset of the dataset (120 pairs for each dataset).

\subsection{Results}
\label{subsec:results}

Our main results are shown in Figure~\ref{fig:results-detection} (see Appendix~\ref{app:full-results} for a full table of results). Overall, we found that false assumption detection is quite challenging, with the best performing models at around 62\% detection accuracy for single-hop (GPT 3.5) and 67\% detection accuracy for multi-hop (GPT-4) using 4-shot in-context examples. There was no substantial benefit of chain-of-thought reasoning on top of the few-shot demonstrations.

\paragraph{Patterns of response bias} Many of the LLMs we tested exhibited clearly biased behavior. For instance, Flan-T5 was heavily biased towards answering ``No'', and PaLM-2 showed a less extreme but similar preference towards ``No'' responses. More interestingly, Llama-2 showed a heavy false assumption bias, most of the times answering that a question had a false assumption regardless of the actual question content.

\paragraph{Effect of search-engine augmentation} The use of FreshPrompt did result in nontrivial gains ($\sim$12.5 percentage points) for the detection task, compared to simple few-shot prompting (Figure~\ref{fig:results-detection}, far right).\footnote{We used the default setting of 5-shot demonstrations and 15 retrievals from \href{https://colab.research.google.com/drive/1YKL3wN1LdaY5Aqlz8IRlag7zzMr7FSTQ}{the official colab}, with minor modifications to the prompt to match the setting for our few-shot evaluation.} However, we note that FreshPrompt, while effective, is a very costly solution---evaluation on a small subset (120 question pairs) from each dataset led to an API cost of $\sim$\$100.

\paragraph{Manual generative QA evaluation} We conducted manual evaluation of generative QA answers for a subset of the questions ($n=200$, 50 questions with/without false assumptions for each of the single-hop and multi-hop datasets) for two models: GPT-4 and Llama-2-70B (4-bit)---the best performing models among the API-access and open-source models we evaluated---using 4-shot in-context demonstrations. Figure~\ref{fig:results-misc} (left) shows the results measured in terms of generative answer acceptability. Here, we observe a large performance gap between single-hop and multi-hop QA performance where multi-hop is substantially more difficult, contra the minor performance gap between single- and multi-hop in the detection task.

\section{Discussion}
\paragraph{Are synthetic, long-tail false assumptions more difficult to detect than naturally occurring ones?} 
To compare our detection results for synthetic questions to performance on naturally occurring questions, we also conducted evaluation of GPT-4 and Llama-2-70B on (QA)$^2$ (Figure~\ref{fig:results-misc}, right). The false assumption bias noted in Section~\ref{subsec:results} was equally present for both natural and synthetic false assumptions in the Llama model. GPT-4 (that did not show trivially skewed answer patterns) showed better performance on naturally occurring questions, suggesting that long-tail questions with false assumptions do impose additional challenges.

\paragraph{Difficulty of generative QA vs. False assumption detection} A surprising observation is that the models performed better on generative QA than on false assumption detection (Figure~\ref{fig:results-misc}, left vs. Figure~\ref{fig:results-detection}, middle).  One possible hypothesis is that the kind of linguistic structure of questions corresponding to the detection task is difficult for LLMs in general, because it requires handling of expressions where there is a question inside a question (e.g., ``Does \{Q\} contain a false assumption?''), or expressions of this effect (e.g., ``\{Q\}: Does this question contain a false assumption?''). This hypothesis is supported by the behavior of Flan-T5-XXL, where it completely fails to handle even trivial questions of this structure, always answering ``No'' to questions such as ``Is \{Q\} a question?''. We hope to investigate this hypothesis more systematically in future work. In general, the difficulty of the detection task also explains the limited benefit of the task decomposition approach that \citet{kim-etal-2023-qa} explored: the detection task, used as a subtask in the task decomposition, may in fact be more difficult for LLMs than the QA task itself. This may be due to the discrepancy in the training data---nested questions are likely more sparsely represented than information-seeking questions.

\section{Conclusion}
We presented Syn-(QA)$^2$, an open-domain synthetic QA dataset including questions with false assumptions generated using similarity-based entity perturbation, covering both single-hop and multi-hop scenarios. Evaluating various LLMs on Syn-(QA)$^2$ shows that false assumptions remain a challenge, and that they are more difficult to detect for synthetic, long-tail questions compared to false assumptions in naturally occurring questions. We hope that Syn-(QA)$^2$ and the generation method we proposed can serve as a useful tool in evaluating the progress towards more robust QA systems.

\section*{Limitations}
This work inherits the limitations of work that involves annotation of false assumptions. As noted in \citet{kim-etal-2023-qa}, claiming that an assumption does not hold often involves claiming that something did \textit{not} happen or some relation does \textit{not} hold, which is intrinsically more challenging to find evidence for, compared to finding evidence for properties or relations that do hold. This is because explicit statements of nonexistence of an entity or a relation are more infrequent due to reporting bias, and in the absence of such explicit statements, the verification of falsehood can only rely on inference from omission. 

Relatedly, the potential risk of the current work is that annotation errors due to the aforementioned difficulty may lead to dissemination of false claims as part of the dataset's public release. If this dataset is used as part of the training data for future models, this may lead to more confident claims about factually false information. We will note these risks as part of the disclaimer that will be released with the dataset, and prevent the dataset from being used in model training to the best of our effort by, for example, releasing the dataset as a password protected archive file.

\bibliography{anthology,custom}

\appendix

\section{Wikidata Properties Used for Similarity Calculation}
\label{app:properties}
\begin{itemize}
    \item  Occupation
    \item Sex
    \item Place of Birth
    \item Educated at
    \item Notable work
    \item Field of work
    \item Member of political party
    \item Award received
\end{itemize}

\section{Full Results}
\label{app:full-results}
The full set of results for our experiments can be found in Tables~\ref{tab:full-results} and \ref{tab:full-results-misc}.
 
\begin{table*}[h]
\centering
\scalebox{1}{
    \begin{tabular}{lcc}
        \toprule
        \textbf{Model} & \textbf{Single-hop} & \textbf{Multi-hop} \\\midrule\midrule
        \multicolumn{3}{c}{\textit{Zero-shot}}\\\midrule
        GPT-4 & 0.50 & 0.62 \\
        GPT-3.5 (\texttt{GPT-3.5-turbo}) & 0.51 &  0.62\\
        Llama-2-70B (4-bit) & 0.40 & 0.37\\
        PaLM-2 (\texttt{text-bison-001}) & 0.29 & 0.27 \\
        Flan-T5-XXL & 0 & 0\\
        \midrule
        \multicolumn{3}{c}{\textit{In-context (4-shot)}}\\\midrule
        GPT-4 & 0.60 & 0.67\\
        GPT-3.5 (\texttt{GPT-3.5-turbo}) & 0.62 & 0.46 \\
        Llama-2-70B (4-bit) & 0.50 & 0.50\\
        PaLM-2 (\texttt{text-bison-001}) & 0.40 & 0.14  \\
        Flan-T5-XXL & 0.49 & 0.48 \\
        \midrule
        \multicolumn{3}{c}{\textit{In-context (4-shot) with CoT}}\\\midrule
        GPT-4 & 0.56 & 0.65\\
        GPT-3.5 (\texttt{GPT-3.5-turbo}) & 0.58 & 0.48 \\
        Llama-2-70B (4-bit) & 0.50 & 0.50\\
        PaLM-2 (\texttt{text-bison-001}) & 0.38 & 0.14\\
        Flan-T5-XXL & 0.50 & 0.50\\\midrule
        \multicolumn{3}{c}{\textit{FreshPrompt}}\\\midrule
        GPT-4 & 0.77 & 0.76 \\\bottomrule
    \end{tabular}
}
\caption{Full results for the false assumption detection task.}
\label{tab:full-results}
\end{table*}

\begin{table*}[h]
\centering
\scalebox{1}{
    \begin{tabular}{lccc}
        \toprule
        \textbf{Model} & \textbf{(QA)$^2$} & \textbf{E2E-QA Single-hop} & \textbf{E2E-QA Multi-hop} \\\midrule\midrule
        \multicolumn{4}{c}{\textit{In-context (4-shot)}}\\\midrule
        GPT-4 & 0.74 & 0.89 & 0.66 \\
        Llama-2-70B (4-bit) & 0.49 & 0.79 & 0.52\\\bottomrule
    \end{tabular}
}
\caption{Full results for (QA)$^2$ and generative QA evaluation.}
\label{tab:full-results-misc}
\end{table*}

\section{Prompts}
The full prompts we used for our experiments are given in Tables~\ref{tab:few-shot-prompts} and \ref{tab:few-shot-cot-prompts}. Refer to \url{https://colab.research.google.com/drive/1YKL3wN1LdaY5Aqlz8IRlag7zzMr7FSTQ} for FreshPrompt prompts and hyperparameters.

\begin{table*}
\begin{tabular}{p{0.9\linewidth}}
    \toprule
    \textbf{4-shot CoT Prompt} \\
    \midrule
     Give answer in the following format. First answer Yes/No on whether the question contains any false premises and then give a rationale behind your answer. Below are few examples\\\\
    
    - Question: ``When did Julius Caesar conquer Australia?'' Answer yes if this question contains any false premises, answer no otherwise\\
      Answer: Yes.\\\\
     
    - Question: ``Where is the Eiffel Tower located?'' Answer yes if this question contains any false premises, answer no otherwise\\
      Answer: No.\\\\
     
    - Question: ``Who is the author of the play 'Romeo and Juliet'?'' Answer yes if this question contains any false premises, answer no otherwise\\
      Answer: No.\\\\
     
    - Question: ``What is the color of the invisible cloak in `Harry Potter'?'' Answer yes if this question contains any false premises, answer no otherwise\\
      Answer: Yes.\\
    \bottomrule
\end{tabular}
\caption{4-shot prompt.}
\label{tab:few-shot-prompts}
\end{table*}

\begin{table*}
\begin{tabular}{p{0.9\linewidth}}
    \toprule
    \textbf{4-shot CoT Prompt} \\
    \midrule
     Give answer in the following format. First answer Yes/No on whether the question contains any false premises and then give a rationale behind your answer. Below are few examples\\\\
    
    - Question: ``When did Julius Caesar conquer Australia?'' Answer yes if this question contains any false premises, answer no otherwise\\
      Answer: Yes.\\
      Rationale: This question contains a false assumption. Julius Caesar never conquered Australia; he lived long before the European discovery of the continent.\\\\
     
    - Question: ``Where is the Eiffel Tower located?'' Answer yes if this question contains any false premises, answer no otherwise\\
      Answer: No.\\
      Rationale: The Eiffel Tower is located in Paris, France\\\\
     
    - Question: ``Who is the author of the play `Romeo and Juliet'?'' Answer yes if this question contains any false premises, answer no otherwise\\
      Answer: No.\\
      Rationale: The author of the play `Romeo and Juliet' is William Shakespeare.\\\\
     
    - Question: ``What is the color of the invisible cloak in `Harry Potter'?'' Answer yes if this question contains any false premises, answer no otherwise\\
      Answer: Yes.\\
      Rationale: This question contains a false assumption. The invisible cloak in `Harry Potter' is described as making the wearer invisible, so it doesn't have a specific visible color when worn.
    \\\bottomrule
\end{tabular}
\caption{4-shot CoT prompt.}
\label{tab:few-shot-cot-prompts}
\end{table*}

\section{License and Terms for Use}
HotpotQA is distributed under CC BY-SA 4.0 License, and (QA)$^2$ is distributed under Apache-2.0 License. We plan to distribute our dataset under Apache-2.0.

\end{document}